\UseRawInputEncoding

\documentclass[10pt,journal,compsoc]{IEEEtran}
\ifCLASSOPTIONcompsoc
  \usepackage[nocompress]{cite}
\else
  \usepackage{cite}
\fi
\ifCLASSINFOpdf
\else
   \usepackage[dvips]{graphicx}
\fi
  \usepackage[caption=false,font=footnotesize,labelfont=sf,textfont=sf]{subfig}
  \usepackage[caption=false,font=footnotesize]{subfig}
\usepackage{multirow}

\begin{document}

%
\title{Multimodal Personal Ear Authentication\\ Using Smartphones}
%
%
%
%

\author{Shunji~Itani, 
		Shunsuke~Kita
        and~Yoshinobu~Kajikawa

\IEEEcompsocitemizethanks{\IEEEcompsocthanksitem 
S.~Itani is with the Graduate School of Science and Engineering of Kansai University, Suita, Japan, \protect\\
E-mail: k938025@kansai-u.ac.jp
\IEEEcompsocthanksitem S.~Kita is with Osaka Research Institute of Industrial Science and Technology, Izumi, Japan
\IEEEcompsocthanksitem Y.~Kajikawa is with Faculty of Engineering Science of Kansai University, Suita, Japan}
}

%
%

\markboth{}%
{Shell \MakeLowercase{\textit{et al.}}: Bare Demo of IEEEtran.cls for Computer Society Journals}
%



\IEEEtitleabstractindextext{%
\begin{abstract}
In recent years, biometric authentication technology for smartphones has become widespread, with the mainstream methods being fingerprint authentication and face recognition. However, fingerprint authentication cannot be used when hands are wet, and face recognition cannot be used when a person is wearing a mask. Therefore, we examine a personal authentication system using the pinna as a new approach for biometric authentication on smartphones. Authentication systems based on the acoustic transfer function of the pinna (PRTF: Pinna Related Transfer Function) have been investigated. However, the authentication accuracy decreases due to the positional fluctuation across each measurement. In this paper, we propose multimodal personal authentication on smartphones using PRTF. The pinna image and positional sensor information are used with the PRTF, and the effectiveness of the authentication method is examined. We demonstrate that the proposed authentication system can compensate for the positional changes in each measurement and improve the robustness.
\end{abstract}

}

\maketitle

\IEEEdisplaynontitleabstractindextext

%
\IEEEpeerreviewmaketitle

\IEEEraisesectionheading{\section{Introduction}\label{sec:introduction}}
%
%
%
%
\IEEEPARstart{R}{ecently}, personal authentication technology has been introduced in online shopping, apartments, companies, and so on. However, since authentication methods that use ID cards and passwords have become the norm, the number of incidents such as spoofing caused by the loss of ID cards and password theft is increasing. Therefore, biometric authentication, a personal authentication technology using an individual's biometric information with robustness and convenience, is drawing attention and interest.\par 
In research on biometric authentication technology, authentication methods using moving images of the pinna, the external part of the ear in humans and other mammals, have been studied~\cite{b1}. Several studies have concluded that the shape of the pinna is complex and differs significantly across individuals~[2-5]. Conventional studies indicate the feasibility of an authentication system using the acoustic transfer functions of the pinna and the ear canal~\cite{b6}. Authentication experiments were conducted using the acoustic transfer function of the pinna: the Pinna Related Transfer Function (PRTF) measured by microphones attached to earphones, headphones, and mobile phones. As a result, the authentication accuracy of the mobile phone was lower than those of the headphones and the earphones. This is because headphones and earphones cover the pinna and the ear canal. An authentication system using canal-type earphones has been proposed to improve the authentication accuracy because the positional fluctuation is low for each measurement~\cite{NEC}. The classification error is shown to be less than 1\%. \par
There have been many studies on personal authentication using earphones, but no research on personal authentication using mobile phones~[8-10]. H. Takemoto et al. showed that the PRTF changes according to the direction in which the measurement signal is emitted to the pinna~\cite{b7}. Furthermore, our previous study suggested that the robustness of the authentication system could be improved by measuring PRTFs from different directions and using them for authentication simultaneously~\cite{b8}. This fact suggests that authentication is possible if the location of the smartphone while measuring is known. In our previous study, we trained a classifier using the positional information obtained from the sensor on the smartphone and used it with the PRTF for simultaneous learning~\cite{b9}. The results showed that the robustness of the personal authentication system is improved by using sensor information to interpolate the location of the smartphone during PRTF measurement. In another previous study~\cite{b11}, it was reported that multimodal personal authentication using voice and pinna images measured from a smartphone can improve the robustness of authentication systems. Therefore, the robustness of personal authentication systems can be improved by using both pinna images and PRTFs for training. \par
In recent years, smartphones have become popular and various biometric authentication methods have been implemented. Fingerprint authentication and face recognition are the most popular biometric authentication methods. However, in the case of fingerprint authentication, the accuracy of collation is affected by the dryness and sebum on the finger. Moreover, the fingerprints can be copied easily, which raises safety concerns. In the case of face recognition, if you are wearing a mask or glasses, if the shape of your face changes due to aging or weight change, or if your face is exposed to light, it may not be recognized. \par
Our conventional system~\cite{b8} uses a speaker and a microphone mounted on the receiver of a smartphone to conduct authentication. In this system, the measurement device uses a time stretched pulse (TSP) signal to measure the PRTF. The amplitude of the frequency response of the PRTF is then input into a classifier for personal authentication. However, it is known that the authentication accuracy is low when only PRTFs are used. In this paper, we propose a new method of multimodal personal authentication using a smartphone. We aim to improve the robustness of our personal authentication system by using a smartphone camera to capture pinna images and simultaneously measure the PRTFs and sensor information. To build this system, we need a dataset that includes the PRTFs, pinna images, and sensor information. In this study, we constructed a mock smartphone equipped with microspeakers, microphones, a camera board, proximity sensors, and 6-axis sensors, and measured the data from nine ears. After preprocessing the data, a Deep Neural Network (DNN) is trained by inputting the data and personal authentication is conducted. Furthermore, the measured input data is visualized in two dimensions using t-Distributed Stochastic Neighbor Embedding (t-SNE), and its effectiveness as a feature of each modal is verified. In addition, by visualizing the values of the middle layer of DNN, we investigate the effectiveness of the multimodal authentication model.

%
%
%
%

\section{Related Work}
\subsection{Authentication Using Pinna Images}

For biometric authentication via pinna images taken at different angles, an algorithm was proposed by Iwakami et al.~\cite{b15} to improve the robustness of the pinna authentication system in case of posture changes. When pinna images acquired by security cameras are matched with images of suspects, it is necessary to consider the difference in the angles of the pinna images because the acquired images are usually taken at different angles from the images of the suspects. The Gabor feature is obtained by a Gabor filter that combines the image and the Gabor function, and the plane corresponding to the uneven shape around the pixel of interest in the image represents the wavelength and direction. This feature is known to be robust to some degree of angular change. Despite the reduction in the number of pixels in the image, the proposed algorithm showed high robustness when seven locations in the pinna region were considered as feature points. This indicates that individuals can be identified based on individual differences in the shape of the pinna. \par
In recent years, with the spread of deep learning and the development of computer vision, Convolutional Neural Network (CNN)-based approaches have been adopted for pinna authentication and have been reported to provide higher accuracy than traditional authentication techniques~[16-18]. CNN requires a large amount of data for training, but the sample size of the dataset available for pinna authentication is limited compared to other fields ~[16, 19]. To combat this, transfer learning can be used; it is effective for training models with a small amount of data~[16, 20]. Transfer learning is a technique of applying knowledge from one domain to another domain and is useful for small amounts of data. Various pre-trained CNN models have been already developed. One of the most famous models that have shown the potential of CNN-based approaches is the AlexNet model. The AlexNet model, introduced by Krizhevsky et al. for object classification in the ImageNet Large Scale Visual Recognition Challenge (ILSVRC), achieved unprecedented performance on ImageNet data, leading to a surge in the use and popularity of CNN-based models~\cite{b22}. In terms of ILSVRC results, the introduction of the VGG16~\cite{b23} model has further improved the authentication rate of ILSVRC~[18, 20]. Recently the two pre-trained models were also used for other tasks and demonstrated high accuracy in pinna authentication.

\begin{figure*}[!t]
\vspace{1.5mm}
  \begin{center}
	\vspace{3.0mm}
       \includegraphics[scale=0.5]{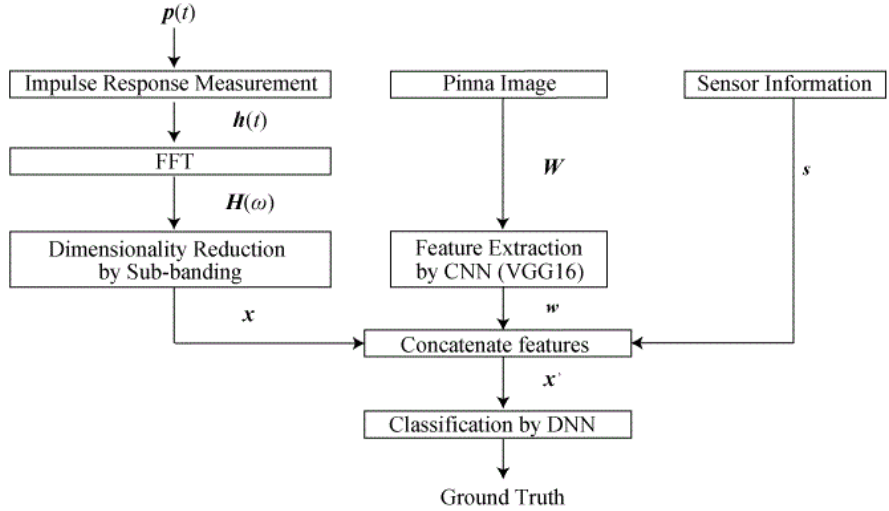}
	\vspace{-2.0mm}
       \caption{Overview of personal authentication using pinna. The input data $\textit{\textbf{x'}}$ are made by combining the feature vectors $\textit{\textbf{x}}$, $\textit{\textbf{w}}$ and the sensor information $\textit{\textbf{x}}$. Standardization is conducted so that the average is 0 and the variance is 1 for the entire training data. Label data is 0 for registrants or 1 for others. }
	\vspace{-5.0mm}
       \label{fig:over}
  \end{center}
\end{figure*}
\vspace{2.5mm}

\subsection{Authentication Using PRTFs}
In a previous study~\cite{NEC}, cosine similarity and SVM were used for authentication, and the effects of earphone attachment and removal were investigated. First, the acoustic transfer function of the ear canal is analyzed by MFCC and the discrimination experiment by cosine similarity is conducted. The results show that the error rate is 1.29\%, confirming the validity of the data as biometric data.
Next, an authentication experiment using SVM was conducted, and the error rate was 0.029\%. This is a high authentication rate for a method using the acoustic transfer function of the ear canal.
Finally, the effect of attaching and detaching the earphones was examined. 
The results show that there are more variations and individual differences in case of subjects who attached or detached the earphones compared to subjects who did not attach or detach them.
These results show that the authentication results were very accurate, but the variation of the data due to the attachment and removal of earphones can be a topic for future studies. \par
S. Yano et al. used four channels of PRTF measured by earphones with four microphones attached at 90 degrees each to reduce the variation in data across cases when the earphones were attached or detached~\cite{NEC2}. As a result, it was found that using four channels of data for training reduced the average error rate by 2-3\%. It also delivered better accuracy compared to the cases in which a single channel PRTF was used. From the comparison of the inter-class variance of the PRTFs obtained from each channel, it was stated that the features obtained by changing the angle of the microphone change. The reasons for the improvement in accuracy are due to the interpolation of variations caused by the increase in angle data and the increase in training data. \par
M. Yasuhara et al. conducted an authentication experiment using inaudible signals for pinna personal authentication using earphones~\cite{NEC3}. It has been reported that inaudible signals can be used for accurate authentication if the position of the earphone relative to the ear canal is constant. On the other hand, when the position of the earphone is changed, the audible signal can be used for more accurate authentication. Therefore, a hybrid system was proposed that uses an audible signal for initial authentication and an inaudible signal for continued authentication after the earphones are worn. \par
In this study, we present a new approach to biometric authentication for smartphones that uses the acoustic transfer function of the pinna for personal authentication.


\section{Proposed Personal Authentication System}
In this chapter, we describe our proposed personal authentication system. We also describe the t-SNE used for data visualization.

\subsection{System overview}
Fig.~\ref{fig:over} shows the overview of the proposed personal authentication system using PRTF. First, the impulse response (PRIR : Pinna Related Impulse Response) $\textit{\textbf{h}}(t)$, pinna image $\textit{\textbf{W}}$, and sensor information $\textit{\textbf{s}}$ (three-dimensional acceleration and the distance between pinna and smartphone) are measured simultaneously time for registrants and non-registrants. 
In the measurement of acoustic inputs, the Time Stretched Pulse (TSP) signal $\textit{\textbf{p}}(t)$ is used. The impulse responses are transformed to the frequency responses $\textit{\textbf{H}}(\omega)$ by FFT. Next, sub-banding is applied by dividing the PRTF into 240 sub-bands and taking the average of the sum of amplitude. The values of the vector after sub-banding are extracted as a feature vector $\textit{\textbf{x}}$. \par
\begin{figure}[t]
\vspace{1.5mm}
  \begin{center}
	\vspace{3.0mm}
       \includegraphics[scale=0.3]{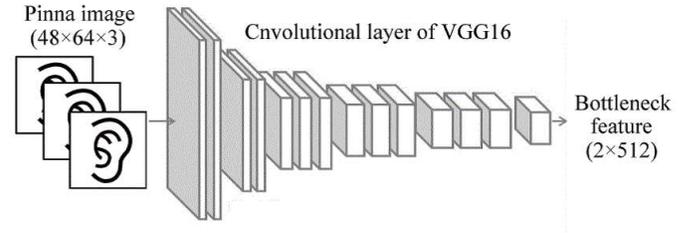}
	\vspace{-2.0mm}
       \caption{Overview of feature extraction of pinna images using VGG16. Pinna images are input into the model from which the fully connected layer of VGG16 is removed and the value of the last convolution layer is extracted as a bottleneck feature.}
	\vspace{-5.0mm}
       \label{fig:vgg16}
  \end{center}
\end{figure}

The acquired pinna image $\textit{\textbf{W}}$ is input into the CNN to extract the bottleneck features. The bottleneck feature is the feature output by the last layer of the convolutional layer in CNN, and it is known that this feature represents the features of the image. Although training a CNN usually requires a large amount of data, transfer learning is used to train the model with a small amount of data. In this study, we use VGG16 as the pre-trained CNN model and use ImageNet-based weights. The values of the feature vectors are extracted as $\textit{\textbf{w}}$ after feature extraction. The extracted bottleneck features are a $2 \times  512$ multidimensional array, but the input must be a one-dimensional array because DNN is used as the classifier. By integrating and flattening each feature map, it is converted into a one-dimensional array with 1024 features. Fig.~\ref{fig:vgg16} shows an overview of feature extraction for a pinna images. \par
The sensor information $\textit{\textbf{s}}$ provides 4-dimensional data that includes 3-axis acceleration and the distance to the pinna when the smartphone is held against the pinna. \par
The input data $\textit{\textbf{x'}}$ are made by combining the feature vector $\textit{\textbf{x}}$, $\textit{\textbf{w}}$ and the sensor information $\textit{\textbf{s}}$. Standardization is conducted so that the average is 0 and the variance is 1 for the entire training data, and label data which is 0 for registrants or 1 for others.
Finally, DNN is used as a two-class classification that categorizes the users as genuine registrants or imposters.

\vspace{2.5mm}

\subsection{t-Distributed Stochastic Neighbor Embedding (t-SNE) }
t-SNE is a machine-learning algorithm for visualization developed by Laurens van der Maaten and Geoffrey Hinton. It is known to be effective for nonlinear data~\cite{tsne} as it maps high-dimensional data sets into two or three dimensions with high probability. 
Multivariate normal distribution is assumed in the high-dimensional space and a student's t-distribution with 1 degree of freedom, taking into account the distances between points, is assumed in the lower dimension so that similar sets are close together and different sets are far apart.

\section{Experiments for Personal Authentication}
In this chapter, we describe the measuring environment and the personal authentication experiment.

\begin{figure}[!t]
 \begin{center}
\vspace{0.5em}
 \subfloat[Front side]{
  \includegraphics[width=0.45\hsize]{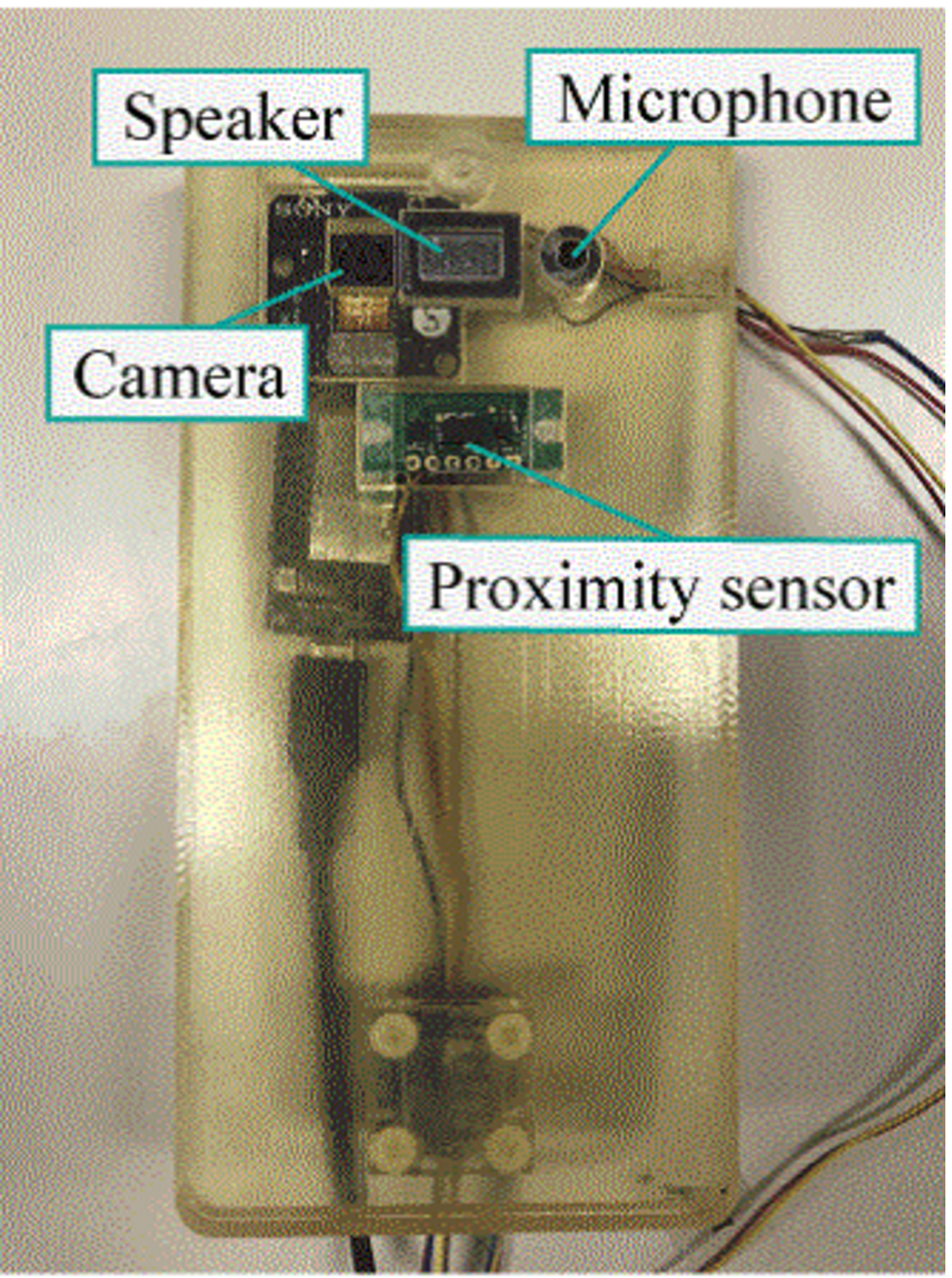}}
   \subfloat[Back side]{
  \includegraphics[width=0.45\hsize]{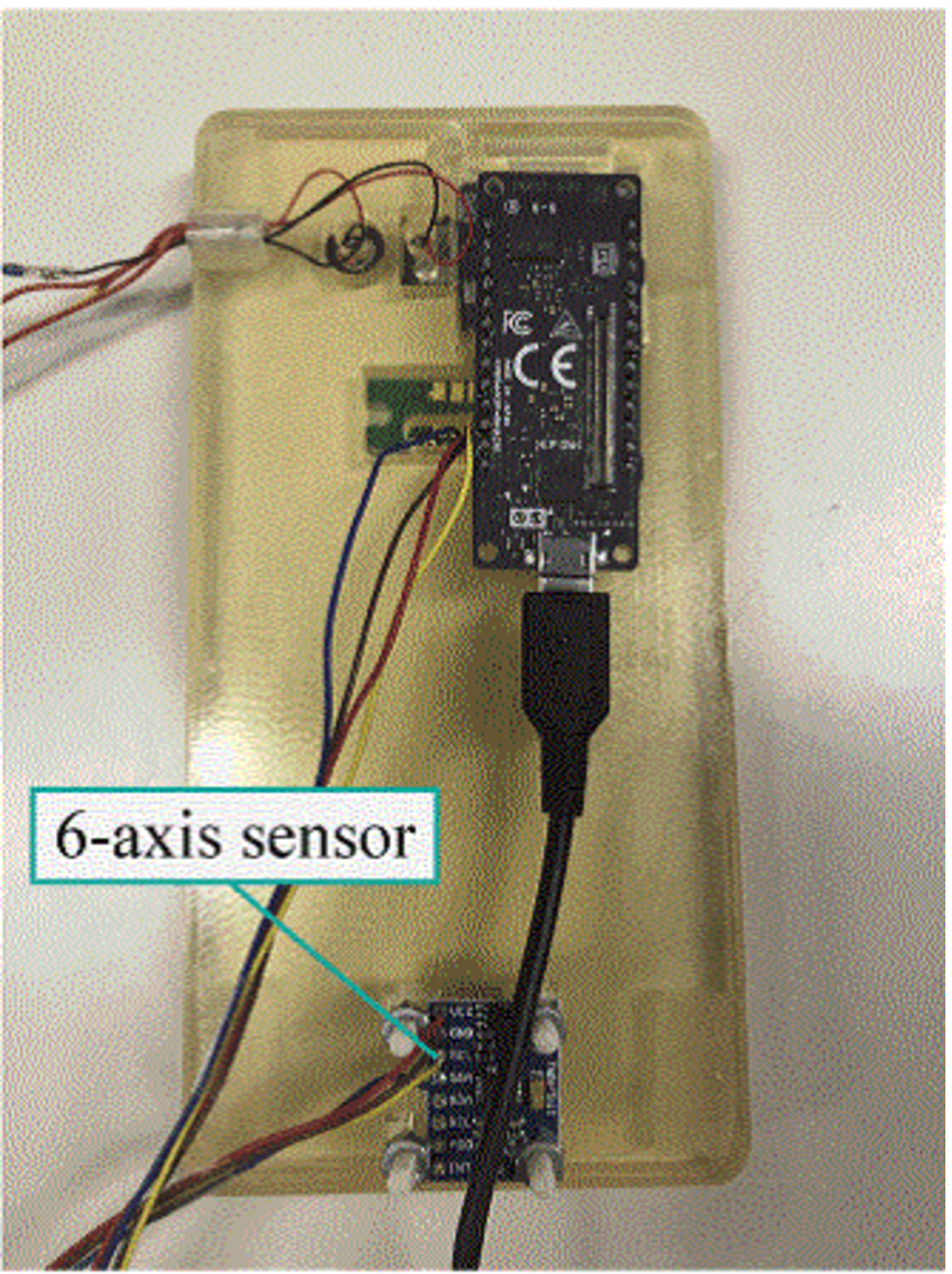}}
   \caption{Handmade mock smartphone used for measurement.}
   \label{mock}
 \end{center}
\end{figure}

\begin{figure}[!t]
\vspace{1.5mm}
  \begin{center}
	\vspace{3.0mm}
       \includegraphics[scale=0.65]{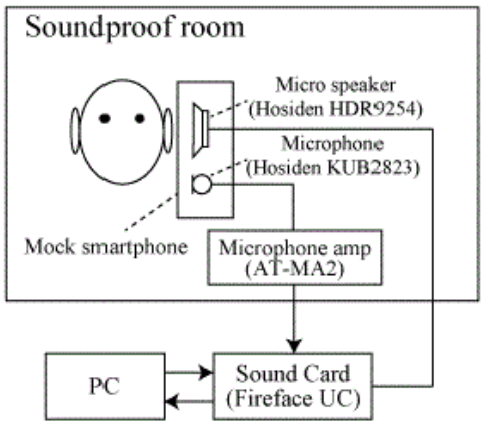}
	\vspace{-2.0mm}
       \caption{Block diagram and configuration of a measurement system for PRTF.}
	\vspace{-5.0mm}
       \label{measure_block}
  \end{center}
\end{figure}
\vspace{2.5mm}

\subsection{Measuring Environment}
Fig.~\ref{mock} shows the front and back of a handmade measurement device like a smartphone. It is equipped with a micro speaker (Hosiden HDR9254), microphones (Hosiden KUB2823), camera board (SONY CXD5602PWBCAM1), proximity sensor (VL6180X), and  6-axis sensor (HiLetgo MPU-6050). Fig.~\ref{measure_block} shows the block diagram of PRTF measurement. The PRTFs are measured by using the handmade measurement device and a personal computer with an external sound card (Fireface UC). The measurements were conducted in a soundproof room (2.9 × 3.1 × 2.1 $m^3$) where the walls are treated with sound-absorbing textile material to suppress reflection. Table~\ref{tb:sim} shows the measurement conditions. 
The subjects were nine men and women in their 20s. Measurements were made in the right ear, 20 times for each person. The measurement device was repositioned every time.

\subsection{Personal Authentication Experiment}
The experimental conditions for personal authentication are shown in Table~\ref{table:cond}. The false rejection rate (FRR), the false acceptance rate (FAR), and the half-total error rate (HTER) are used as the evaluation criteria. In the experiment, one ear is selected in order from Ear 1 to Ear 9 as the registrant's ear and the rest are used as ears for imposters. This process is repeated so that all the ears become the registrants, and the average of the authentication results is obtained. \par
In this study, we conduct three different verifications. First, we perform an authentication experiment using a single modal representing the three modals. Next, we discuss the case where two or more modals are combined and input to a DNN. Finally, we examine authentication experiments when two or more modals are input to a DNN with different input layers.

\subsubsection{Case 1: Authentication with a Single Modal}
\begin{figure}[!t]
 \begin{center}
\vspace{0.5em}
 \subfloat[Acoustic model]{
  \includegraphics[width=0.3\hsize]{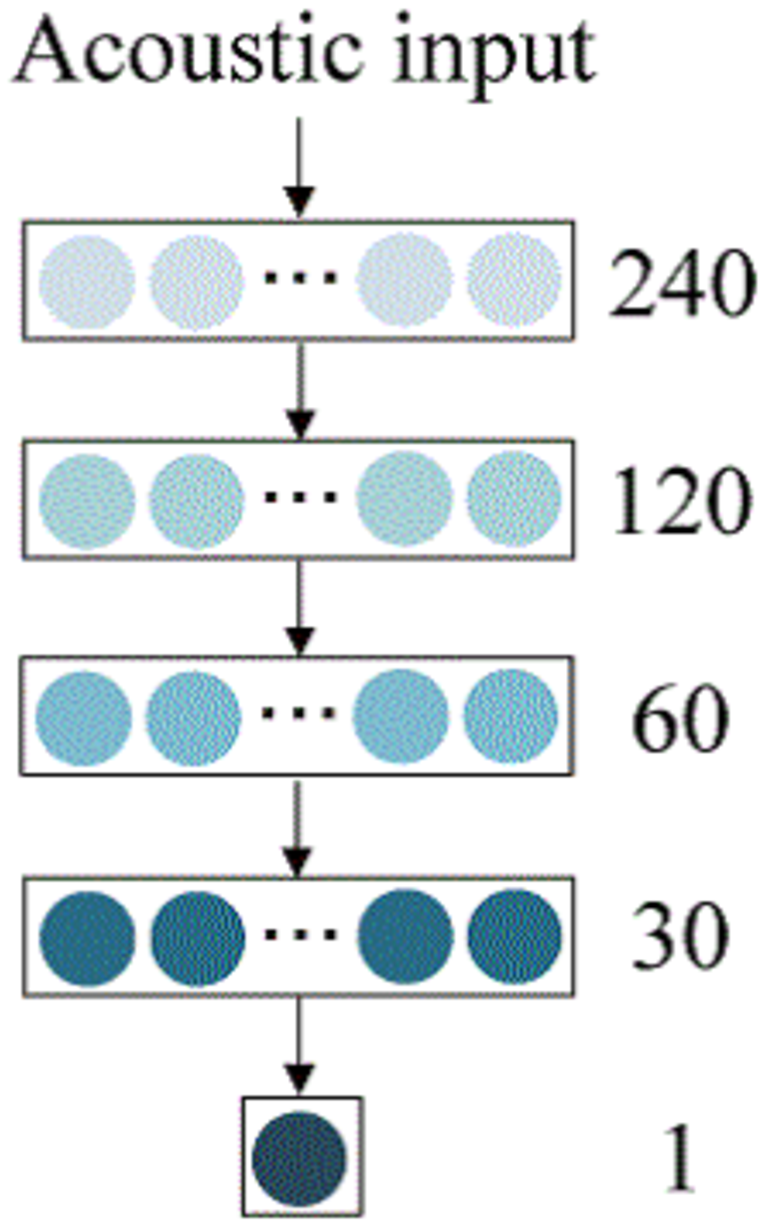}}
  \hspace{5pt}
   \subfloat[Image model]{
  \includegraphics[width=0.3\hsize]{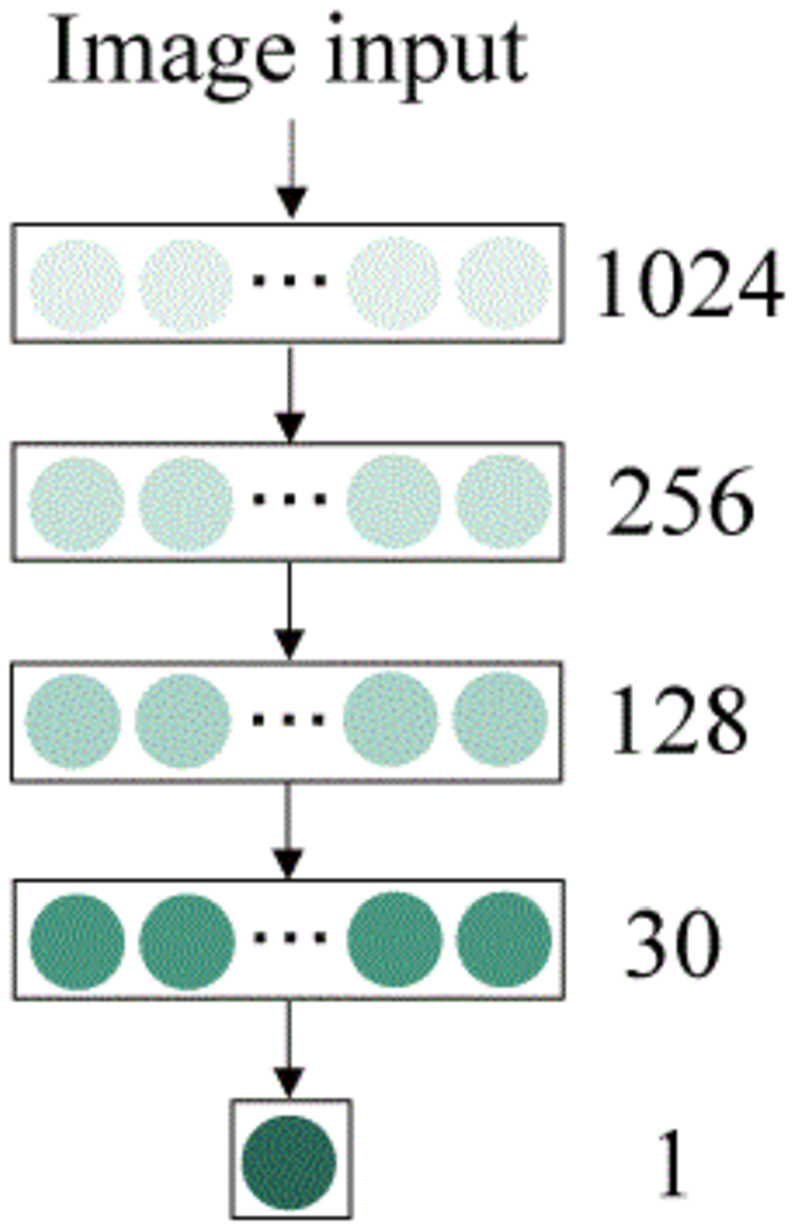}}
	\subfloat[Sensor model]{
	\hspace{5pt}
  \includegraphics[width=0.3\hsize]{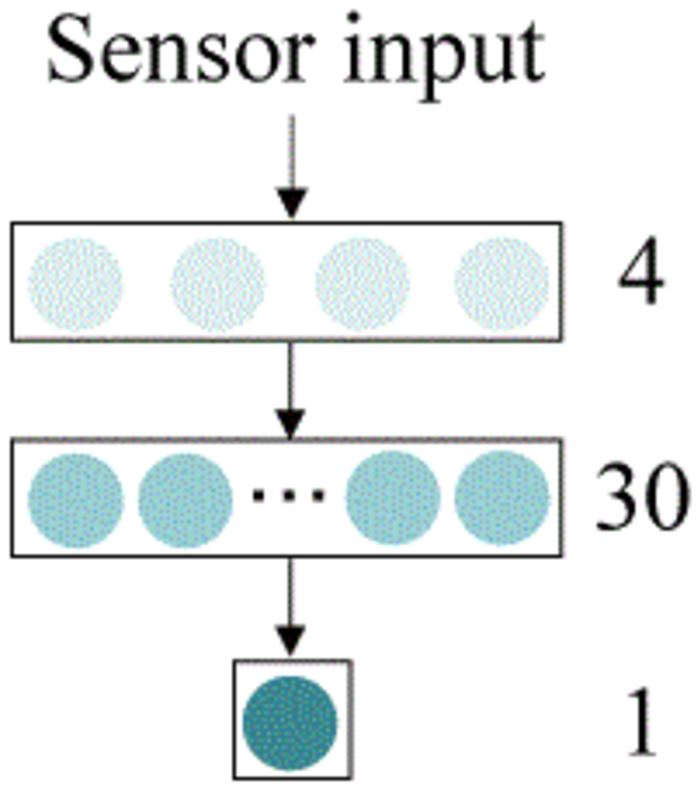}}
   \caption{Model in case of single modal (Case 1). (a): Acoustic model. (b): Image model. (c): Sensor model.}
   \label{single_model}
 \end{center}
\end{figure}

We examine the personal authentication experiment in the case where the PRTF, pinna image, and sensor information are inputted individually to the DNN. 
The number of DNN units is shown in each row representing combinations of acoustics, image, and sensor in Table~\ref{table:model_struc}. 
The DNN with each modal as input data is shown in Fig.~\ref{single_model}.

\subsubsection{Case 2: Authentication with Combined Input Data}
We examine the personal authentication experiment when the input data is combined and input to the DNN. We construct a DNN with four combinations of the three modals: PRTF and sensor information, pinna image and sensor information, PRTF and pinna image, and PRTF, pinna image and sensor information. 
The number of DNN units is shown in each row as acoustics + sensor, image + sensor, acoustic + image and acoustic + image + sensor in Table~\ref{table:model_struc}. 
For example, DNN with PRTF and pinna image as input data is shown in Fig.~\ref{mm_model} (a).

\subsubsection{Case 3: Authentication without Combining Input Data}
\begin{figure}[!t]
 \begin{center}
\vspace{0.5em}
 \subfloat[Combined input model]{
  \includegraphics[width=0.5\hsize]{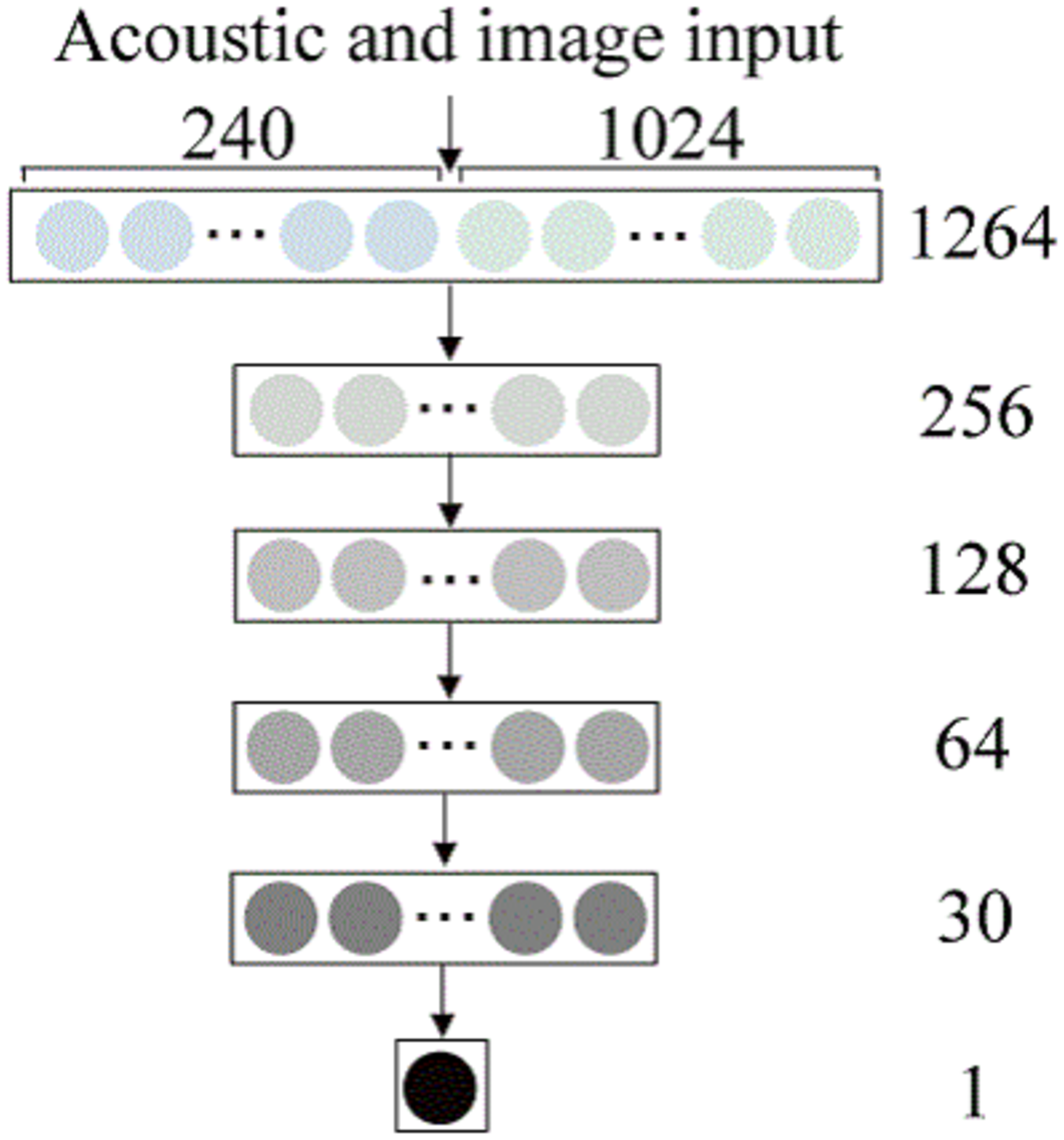}}
   \subfloat[Without combining input model]{
  \includegraphics[width=0.5\hsize]{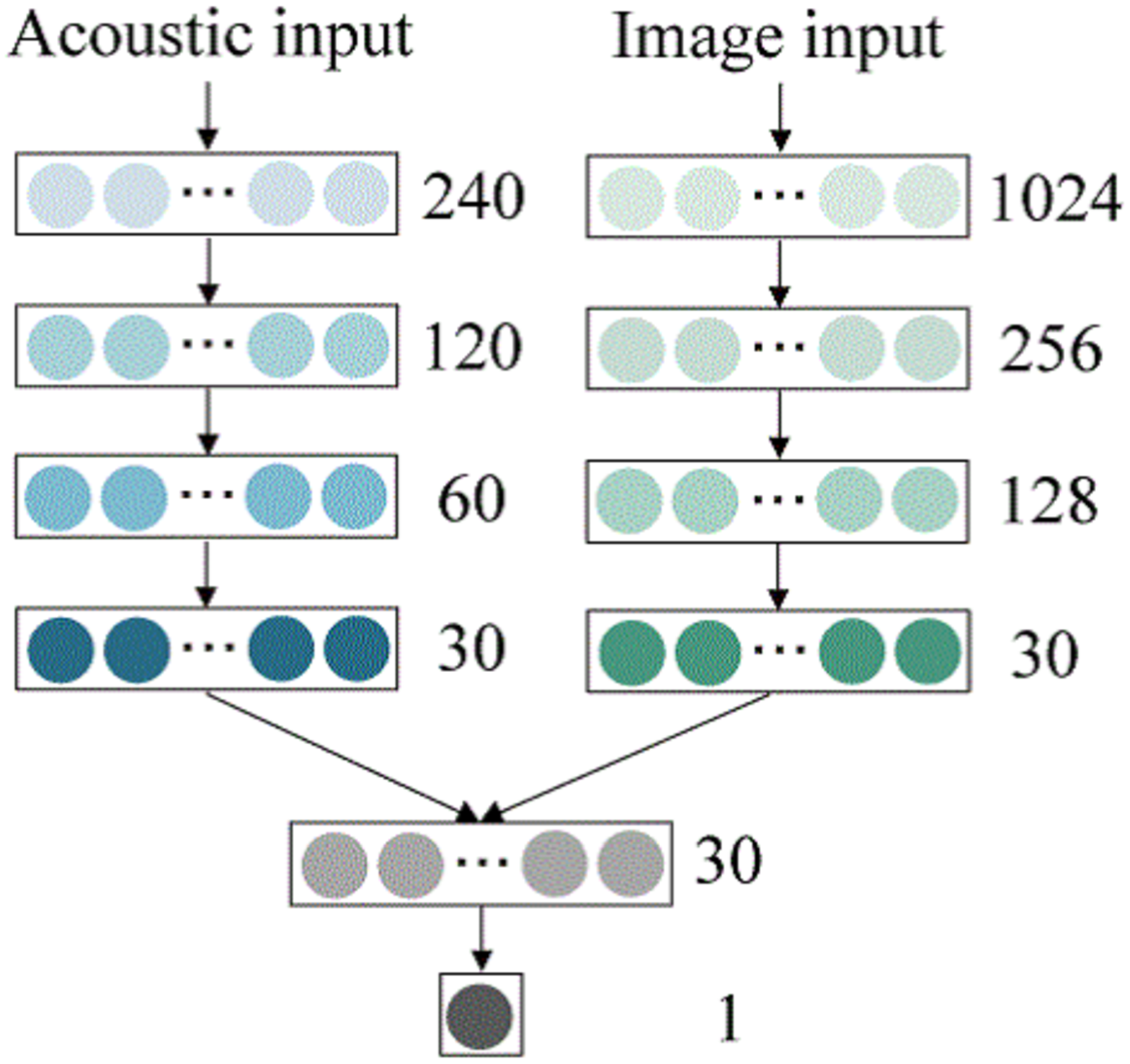}}
   \caption{Model in case of multimodal. (a): A model which combines two modals and inputs them to the DNN  (Case 2). (b): A model which inputs two modals into a DNN with separate input layers  (Case 3).}
   \label{mm_model}
 \end{center}
\end{figure}

To eliminate the imbalance in the dimensions of features of the input data, we examine the personal authentication experiment when data is input into the different input layers of DNN. Note that the input data is combined in section 4.2.2, but here it is input without combining. As in 4.2.2, the DNN is constructed with four combinations. 
The number of DNN units is shown in each row of acoustics $\times$ sensor, image $\times$ sensor, acoustic $\times$ imaage and acoustic $\times$ image $\times$ sensor in the Table~\ref{table:model_struc}. 
For example, Fig.~\ref{mm_model} (b) shows DNN with PRTF and pinna image as input data. As shown in the figure, after reducing the dimensionality of the input data in each network, the authentication result output is obtained from a middle layer that merges the outputs from each network.

\begin{table}[!t]
\caption{Measurement conditions}
\begin{center}
\begin{tabular}{l r}
\hline
        	Number of subjects		& 9  \\
		Input signal			& TSP signal \\
		Sampling rate	& 48000 Hz\\
		Length of TSP signal			& 65536 \\
		Speaker input voltage	& 0.149 V \\
		Sound pressure of speaker (SPL)	& 88.4 dB \\
		Camera input voltage & 3.3 V  \\
		6-axis sensor input voltage & 3.3 V  \\

\hline
\label{tb:sim}
\end{tabular}
\end{center}
\end{table}

\begin{table}[!t]
	\caption{Experimental conditions}
        \vspace{-6mm}
        \label{table:cond}
        \begin{center}
        \scalebox{1.0}{
		\begin{tabular}{l r}
    	    		\hline
    	    		Registrant training data			 	& 12					\\
			Imposter training data              		& 96					\\
			Registrant test data					& 8					\\
			Imposter test data  					& 64					\\
              Label for registrant 					& 0			\\
              Label for imposters 					& 1		\\
			Dimensions of acoustic input			& 240		\\
			Dimensions of Image input			& 1024	\\
			Dimensions of sensor input			& 4		\\
			Activation function (Middle layer)    &ReLU		\\
			Activation function (Output layer)	&Sigmoid		\\
              Optimization algorithm					&Adam	\\
			Dropout rate	  							&50\%		\\ 
			Evaluation index						&FRR, FAR, HTER\\
                        \hline
		\end{tabular}
		}
        \end{center}
\end{table}

\begin{table}[!t]
	\caption{DNN structure}
        \vspace{-6mm}
        \label{table:model_struc}
        \begin{center}
        \scalebox{1.0}{
  \begin{tabular}{c|c|l}
    \hline
    　   & Model　& Number of units   \\ \hline
 		    \multirow{3}{*}{Case 1} & Acoustic   						& [240, 120, 60, 30, 1] \\
 		    &Image   							& [1024, 256, 128, 30, 1] \\
 		    &Sensor  							& [4, 30, 1] \\ \hline
 		    \multirow{4}{*}{Case 2} &Acoustic + Sensor   			& [244, 120, 60, 30, 1] \\
 		    &Image + Sensor  				& [1028, 256, 128, 64, 30, 1]  \\
 		    &Acoustic + Image   				& [1264, 256, 128, 64, 30, 1] \\
 		    &Acoustic + Image + Sensor  	& [1268, 256, 128, 64, 30, 1]  \\ \hline
			\multirow{9}{*}{Case 3} &\multirow{2}{*}{Acoustic $\times$ Sensor} 	& [240, 120, 60, 30,　\multirow{2}{*}{30, 1]} \\ 
												&& [4, 30 \\
			&\multirow{2}{*}{Image $\times$ Sensor} 		& [1024, 256, 128, 30, \multirow{2}{*}{30, 1]} \\
												&& [4, 30, \\ 
			&\multirow{2}{*}{Acoustic $\times$ Image} 	& [240, 120, 60, 30, 　\multirow{2}{*}{30, 1]} \\ 
												&& [1024, 256, 128, 30, \\
			&\multirow{3}{*}{Acoustic $\times$ Image $\times$ Sensor} & [240, 120, 60, 30, 　\multirow{3}{*}{30, 1]}\\
												&& [1024, 256, 128, 30, \\
												&& [4, 30, \\ \hline 
  \end{tabular}
		}
        \end{center}
\end{table}


\section{Results and Discussion}
In this section, we first confirm the measured data and examine its effectiveness as a feature of each modal by visualizing it in two dimensions using t-SNE. 
In section 5.2, the authentication result of case 1 is shown and the output of the middle layer is visualized. In section 5.3, the authentication result of case 2 is shown and the output of the middle layer is visualized. In section 5.4, the authentication result of case 3 is shown and the output of the middle layer is visualized. As a result, in case 3, FAR is improved in most models and HTER is improved in all models.
From the visualization of the output of the middle layer when the authentication accuracy is high, we can see that the registrant data is clustered at the edge of the distribution so that the registrant and impostor data are separated.

\subsection{Measurement results}
\begin{figure}[t]
\vspace{1.5mm}
  \begin{center}
	\vspace{3.0mm}
       \includegraphics[scale=0.28]{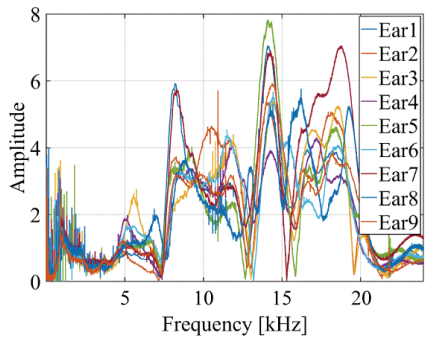}
	\vspace{-2.0mm}
       \caption{PRTFs measured from 9 subjects. }
	\vspace{-5.0mm}
       \label{prtf_9}
  \end{center}
\end{figure}

Fig.~\ref{prtf_9} shows the PRTFs of nine subjects. It can be seen that the PRTFs have similar characteristics up to 5 kHz, but the peaks and dips are very different at frequencies above 5 kHz. This is thought to be due to the change in resonance around the pinna and the change in reflected sound caused by the differences in the shape of the pinna. Therefore, it was shown that individual differences in the pinna shape would result in differences in PRTF. \par
\begin{figure}[t]
\vspace{1.5mm}
  \begin{center}
	\vspace{3.0mm}
       \includegraphics[scale=0.35]{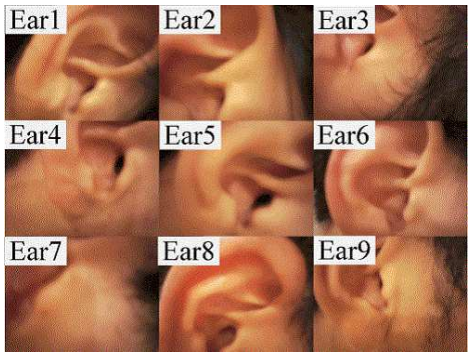}
	\vspace{-2.0mm}
       \caption{Pinna images measured from 9 subjects}
	\vspace{-5.0mm}
       \label{img_9}
  \end{center}
\end{figure}

Fig.~\ref{img_9} shows the nine pinna images from Ear 1 to Ear 9. It can be seen that some images show the entire ear and others only a part of the ear, because different people hold their smartphones in different ways, resulting in variations in distance and angle when taking pictures. Because of the individual differences in the angle and distance in pinna imaging, it is an accepted practice used identifying individuals. \par
\begin{figure*}[!t]
 \begin{center}
\vspace{0.5em}
 \subfloat[Acoustic input]{
  \includegraphics[width=0.35\hsize]{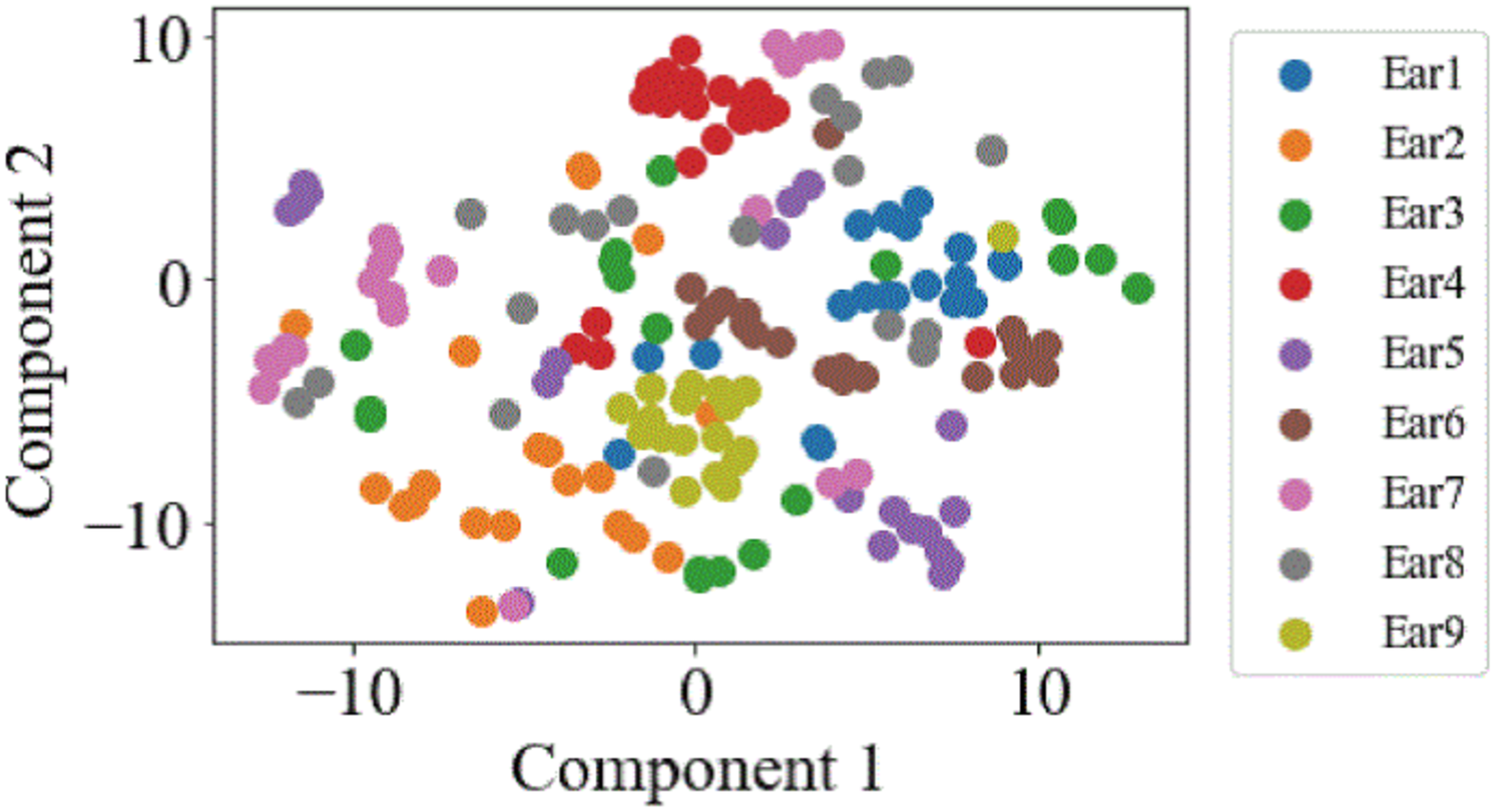}}
   \subfloat[Image input]{
  \includegraphics[width=0.35\hsize]{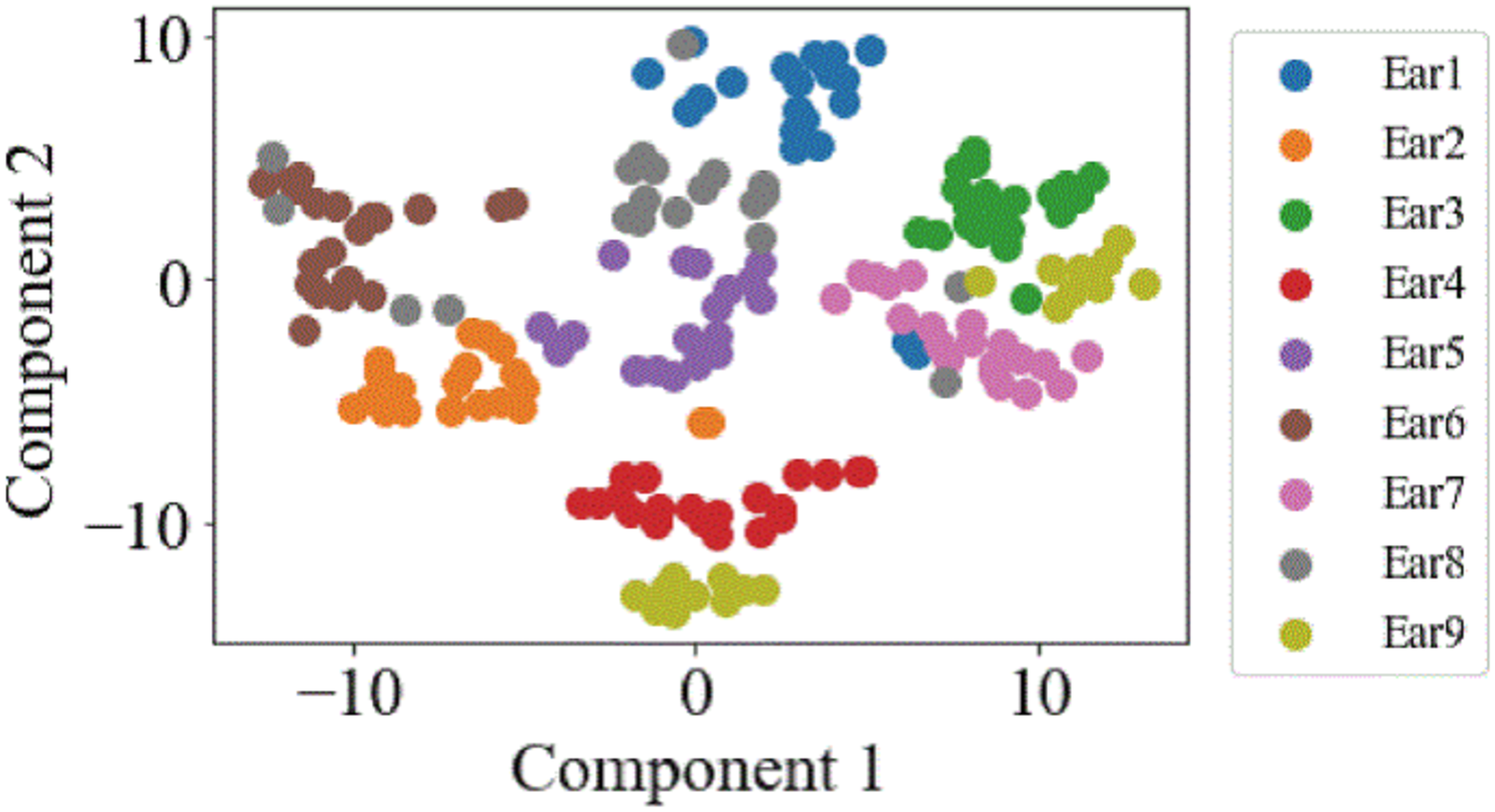}}
	\subfloat[Sensor input]{
  \includegraphics[width=0.35\hsize]{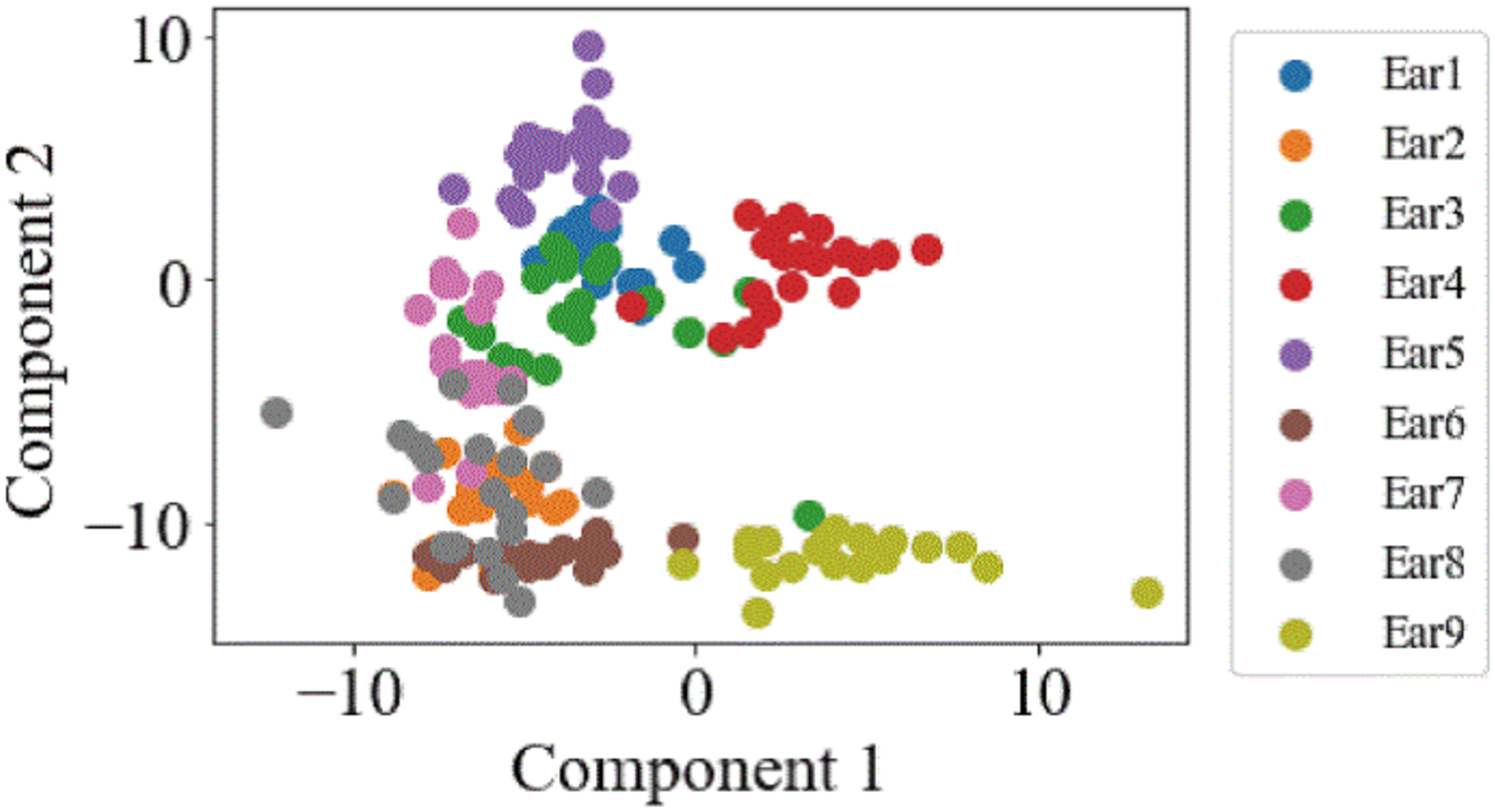}}
   \caption{The results of applying t-SNE for each input data. (a): Acoustic input after dimensionality reduction by sub-banding. (b): Image input after feature extraction by VGG16. (c): Sensor input.}
   \label{input_tsne}
 \end{center}
\end{figure*}

Fig.~\ref{input_tsne} (a) shows the visualization of the PRTFs of Ear 1 to Ear 9 by using t-SNE.
Fig.~\ref{input_tsne} (a) shows that clusters are formed at Ear 4 and Ear 9 but the distribution across the other pinnae is widespread. This is due to the large variation in the position of the smartphone when the subject repositioned it for each measurement, causing the PRTF to fluctuate even for the same pinna. \par
Fig.~\ref{input_tsne} (b) shows the visualization of the pinna images of Ear 1 to Ear 9. t-SNE is used for the preprocessed pinna images as described in \textbf{3.2}. From Fig.~\ref{input_tsne} (b), it can be seen that the pinna images are effective as biometric information because clusters are formed for each ear. \par
Fig.~\ref{input_tsne} (c) shows the visualization of the sensor information of Ear 1 to Ear 9. From Fig.~\ref{input_tsne} (c) clusters are formed in each ear. This suggests that there are individual differences in the way people hold their smartphones, and this can be a feature for identifying individuals. However, it can be seen that the distributions of different ears overlap each other. This is because the angles of the smartphone and the distance to the pinna are within a limited range when the smartphone is held against the pinna. 

\vspace{2.5mm}

%

\begin{figure*}[!t]
 \begin{center}
\vspace{0.5em}
 \subfloat[Acoustic input]{
  \includegraphics[width=0.33\hsize]{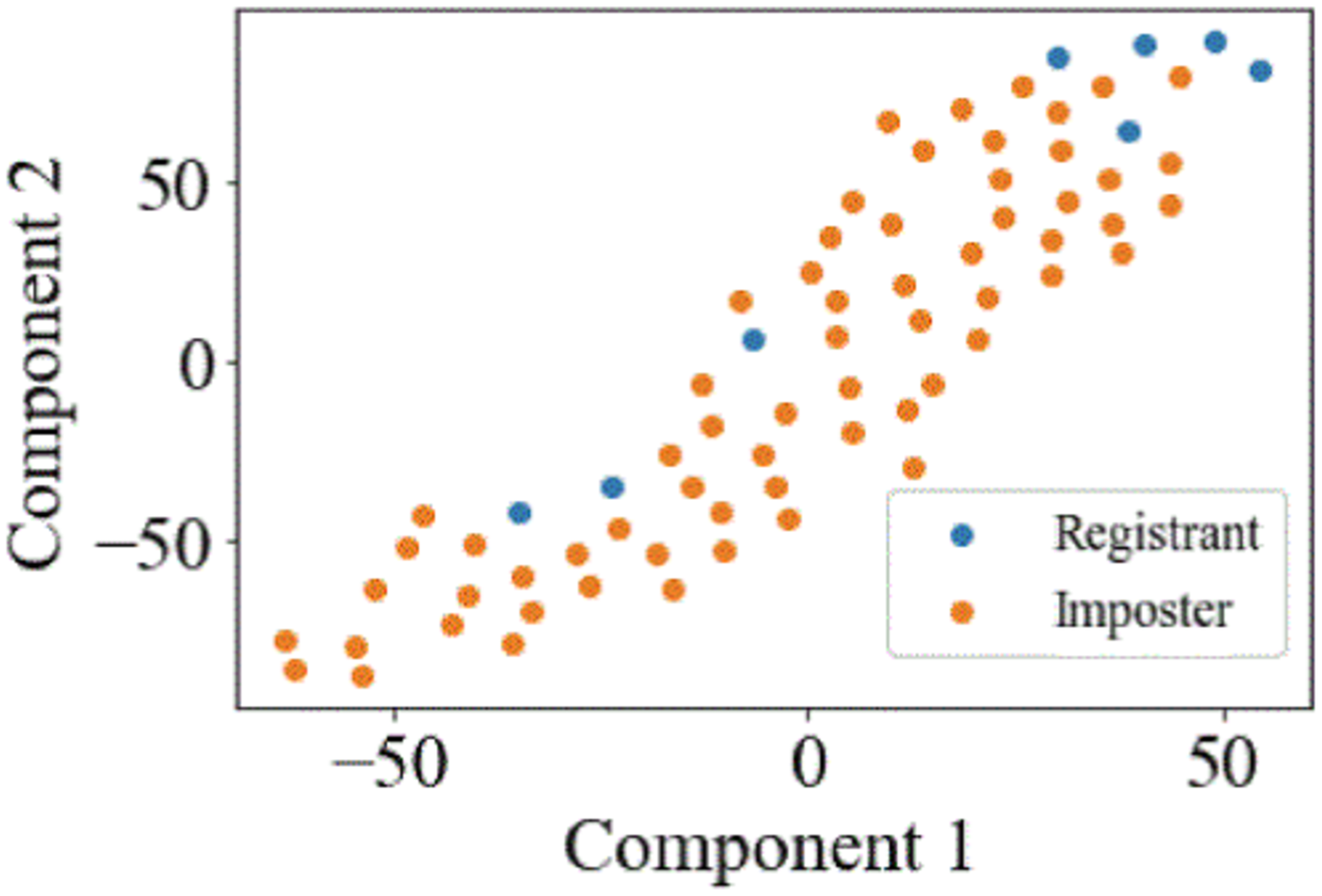}}
   \subfloat[Image input]{
  \includegraphics[width=0.33\hsize]{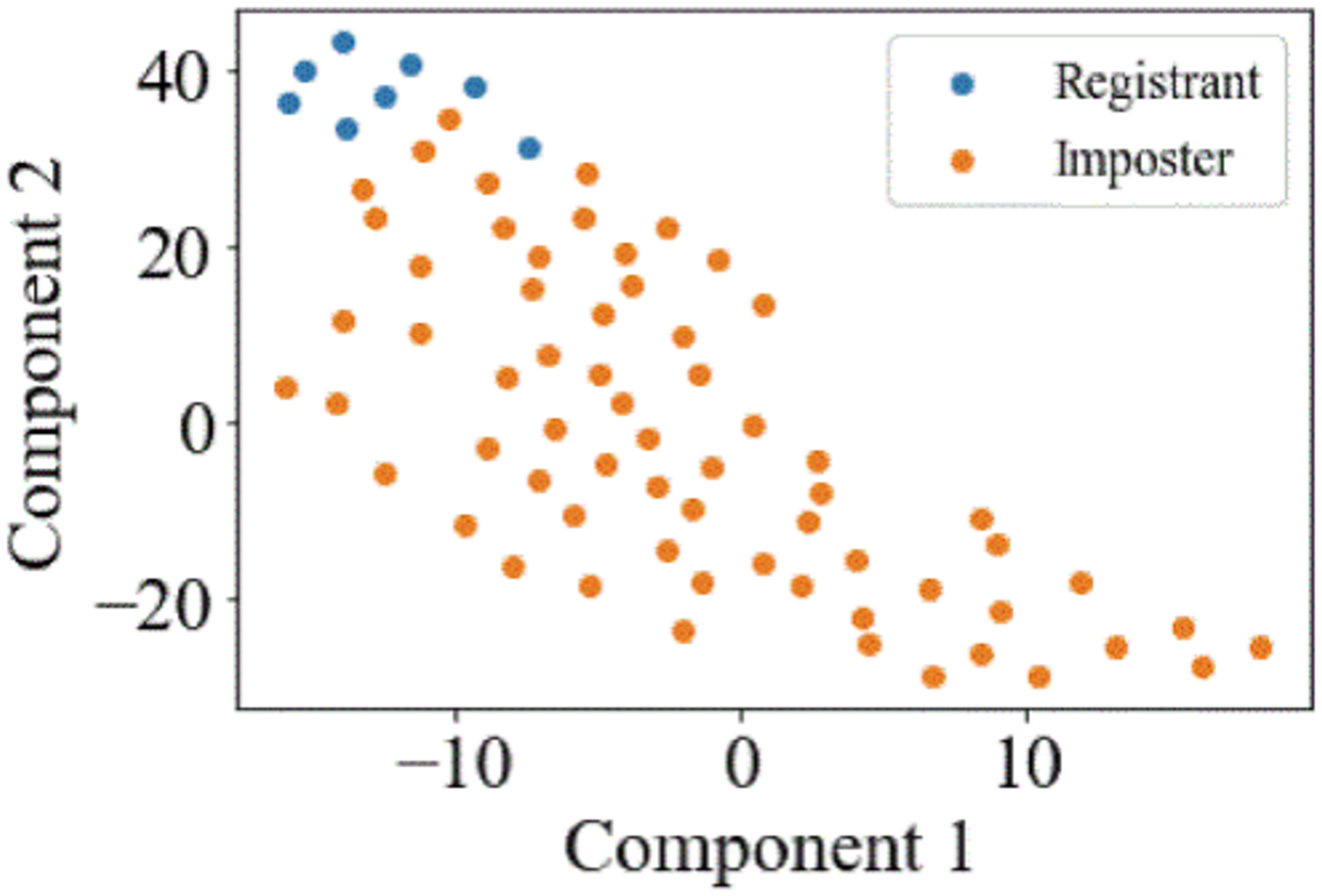}}
	\subfloat[Sensor input]{
  \includegraphics[width=0.33\hsize]{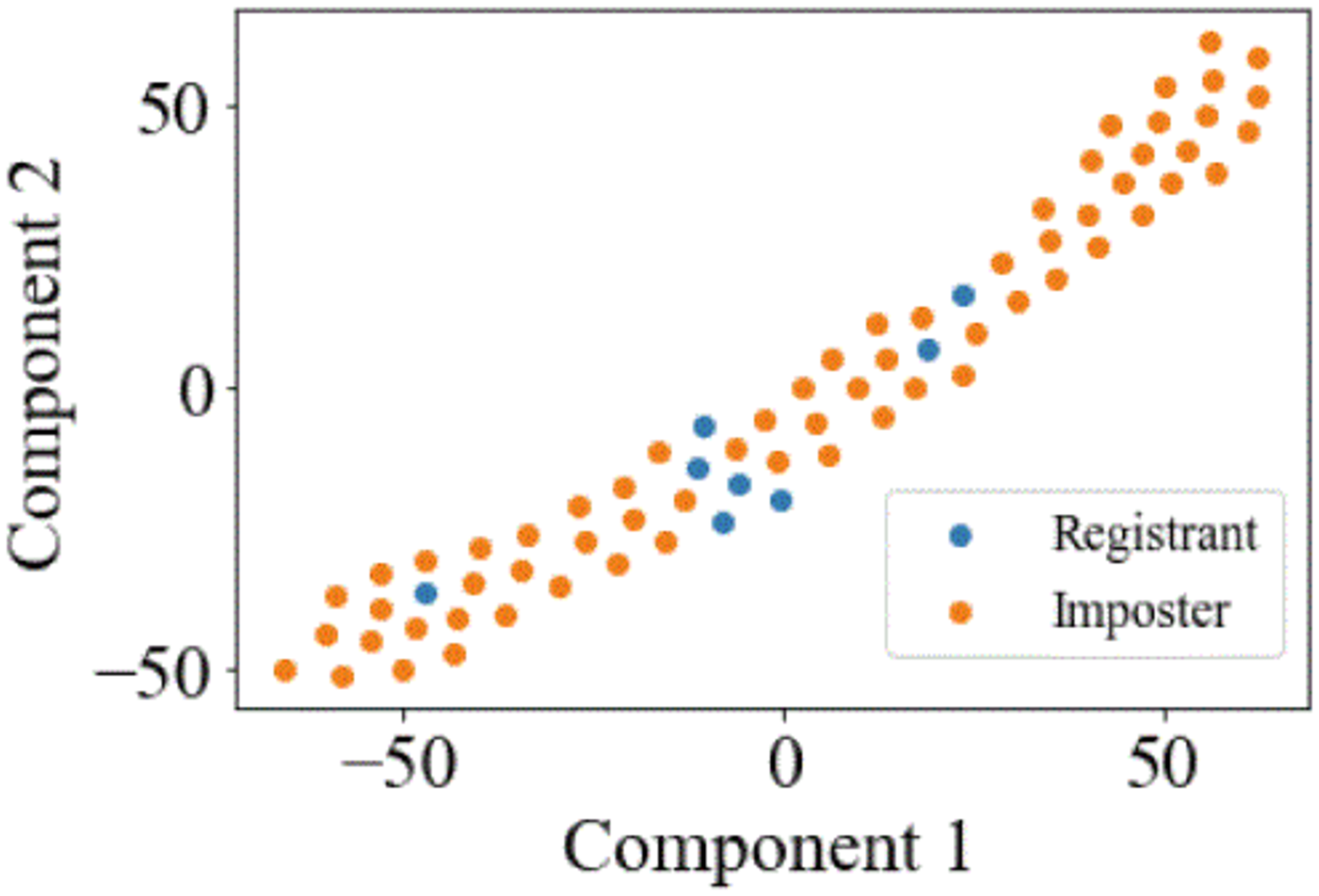}}
   \caption{The results of visualizing the output from the middle layer in the model using single modal. (a): Acoustic model. (b): Image model. (c): Sensor model.}
   \label{sm_tsne}
 \end{center}
\end{figure*}

\subsection{Results of Case 1: Authentication in a single modal }
Case 1 in Table~\ref{result1} shows the authentication results when a single modal is input to the DNN individually. The acoustic model results in a high FAR. The PRTFs of the registrant's data are close to the PRTFs of the imposter's data due to the positional variation during the measurement. In the image model, the FRR and FAR are both low and the authentication accuracy is higher than those of the acoustic model and sensor model. The feature extraction using CNN was robust to scaling and rotation and effective for authentication. The sensor model results in a high FRR. This is because the angle of the smartphone and the distance to the ear can be measured within a certain limited range, so that the same positional information is measured and thus incorrectly classified. \par
After obtaining the output values of the middle layer are obtained, the distribution is confirmed by visualizing them in two dimensions because the values of the middle layer are considered to represent the features for classifications. Each model was trained with Ear 5 as the registrant and the output from the middle layer were visualized when test data were introduced. \par
The results of visualizing the middle layer's output of the acoustic model are shown in Fig.~\ref{sm_tsne} (a). Registrant data is considered misclassified because it is mixed with imposter data. Subsequently, the results of visualizing the middle layer's output of the image model are shown in Fig.~\ref{sm_tsne} (b). Registrant data is gathered at the edge and separated from the imposter data. Hence, it is considered that the authentication accuracy is high. \par
Finally, the results of visualizing the middle layer's output of the sensor model are shown in Fig.~\ref{sm_tsne} (c). Registrant data is considered misclassified because it was mixed with imposter data. The results of the above visualization demonstrated that the image model has the highest authentication accuracy.

\begin{figure}[!t]
\vspace{1.5mm}
  \begin{center}
	\vspace{3.0mm}
       \includegraphics[scale=0.33]{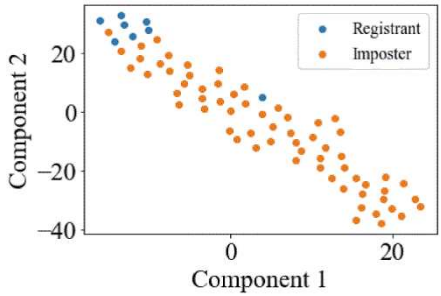}
	\vspace{-2.0mm}
       \caption{Result of visualizing the output from the middle layer of the acoustic + image + sensor model. }
	\vspace{-5.0mm}
       \label{mm1_tsne}
  \end{center}
\end{figure}
\vspace{2.5mm}

\begin{figure}[!t]
\vspace{1.5mm}
  \begin{center}
	\vspace{3.0mm}
       \includegraphics[scale=0.33]{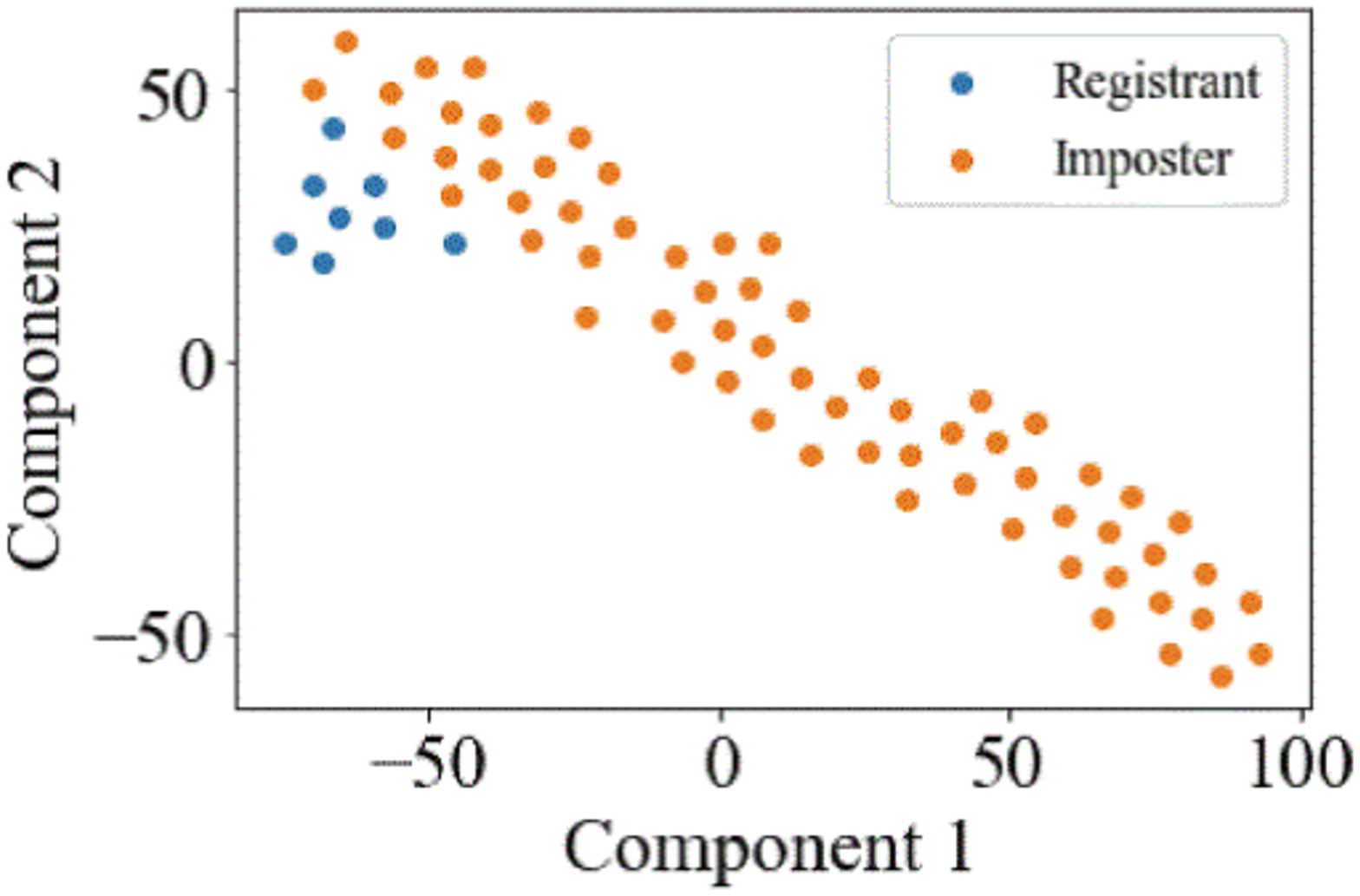}
	\vspace{-2.0mm}
       \caption{Result of visualizing the output from the middle layer of the acoustic $\times$ image $\times$ sensor model. }
	\vspace{-5.0mm}
       \label{mm2_tsne}
  \end{center}
\end{figure}
\vspace{2.5mm}

\begin{table}[!t]
\begin{center}
\caption{Authentication results}
\label{result1}
  \begin{tabular}{c|c|clclcl}
    \hline
    　    & Model　& FRR &FAR &HTER    \\ \hline

 		  \multirow{3}{*}{Case 1} & Acoustic   						& 1.9\% & 16.7\% & 9.3 \%  \\
 		   & Image   							& 1.0\% & 4.2\% & 2.6\%  \\
 		   & Sensor  							& 14.6\% & 5.6\% & 10.1\% \\ \hline
 		  \multirow{4}{*}{Case 2} & Acoustic + Sensor   			& 0.9\% & 19.4\% & 10.2\% \\
 		   & Image + Sensor  				& 1.0\% & 4.2\% & 2.6\%  \\
 		   & Acoustic + Image   				& 0.2\% & 8.3\% & 4.3\%  \\
 		   & Acoustic + Image + Sensor  	& 0.9\% & 6.9\% & 3.9\%  \\ \hline
		  \multirow{4}{*}{Case 3}  & Acoustic $\times$ Sensor    			& 0.4\% & 13.9\% & 8.2\% \\
 		   & Image $\times$ Sensor  				& 0.7\% & 4.2\% & 2.6\%  \\
 		   & Acoustic $\times$ Image   				& 0.5\% & 2.8\% & 1.6\%  \\
 		   & Acoustic $\times$ Image $\times$ Sensor  	& 0.5\% & 2.8\% & 1.6\%  \\ \hline

  \end{tabular}
\end{center}
\end{table}

\subsection{Results of Case 2: authentication with Combined Input Data}
Case 2 in Table~\ref{result1} shows the authentication results when the combined input data are used. Table~\ref{result1} shows that the image + sensor model has the same authentication accuracy as the image model. This is because the number of features of the sensor information is four while the number of features of the pinna image is 1024, which means that the influence of sensor information on authentication is less. The acoustic + image model and acoustic + image + sensor model can improve the authentication accuracy more than the acoustic model, but the authentication accuracy is lower than the image model. This indicates that when features with low authentication accuracy are combined with the image features, the authentication accuracy of the model decreases. \par
The results of visualizing the middle layer's output of the acoustic + image + sensor model are shown in Fig.~\ref{mm1_tsne}. 
One of the registrant data is mixed with the distribution of imposters due to combining acoustic and sensor features with low authentication accuracy with an image feature. Hence, it is thought that the authentication accuracy was lower than that of the image model.

\subsection{Results of Case 3: Authentication without Combining Input Data}
Case 3 in Table~\ref{result1} shows the authentication results when the input data is not combined and input to DNN with different input layers. As we compare Case 2 and Case 3, we can see that although some models show a decrease in FRR when the input data is not combined, all models, except the acoustic + image models, show a significant improvement in FAR  and HTER. 
Therefore, it is considered that the model structure shown in Fig.~\ref{mm_model} (b) is effective for multimodal data. That is, an input layer is separately created for each model, and then the outputs reducing the dimensions in each network are merged via the middle layer. \par
The results of visualizing the middle layer's output of the acoustic $\times$ image $\times$ sensor model are shown in Fig.~\ref{mm1_tsne}. 
Registrant data is gathered at the edge and separated from the imposter data. Furthermore, the registrant data and the imposter data do not overlap each other more than the distribution in the image model. From this, it is considered that the authentication accuracy is higher than that of the image model. From the above results, the acoustic $\times$ image $\times$ sensor model is found to be the most effective for personal authentication based on pinna using smartphones.

\section{Conclusion}
In this paper, we proposed a new biometric authentication system using a smartphone and investigated its effectiveness. The proposed system can improve the robustness of personal authentication systems by using the acoustic transfer function of the pinna measured by the smartphone, a pinna image taken by the smartphone, and the location information obtained from the sensor for learning.
The results showed that the image model has the highest authentication accuracy for authentication by single modal. This can be attributed to the effectiveness of the CNN-based image feature extraction method. When the features of two or more modals were combined and input to the DNN, the authentication accuracy was lower than when only image features were used. It was found when the image features are combined with the features with low authentication accuracy, the authentication accuracy of the model decreases. \par
Subsequently, to eliminate the imbalance in the number of features in the input data, we examined the personal authentication experiments when the input data are separately applied to DNN with different input layers without combining the input data. The results show that the authentication accuracy can be improved in all models. In the case of multimodal networks, it is effective to create an independent input layer for each modal, reduce the dimensionality with each network, and then output from the merged middle layer.

For future studies, we are trying to find a way to extract the features of PRTFs. In this study, the features of the pinna image had a greater impact on authentication than the other features. This can be attributed to the effectiveness of the feature extraction method for pinna images using CNN's transfer learning. On the other hand, the PRTF was sub-banded by averaging the amplitude values. Although this method is effective in reducing the number of dimensions, it is insufficient as a method for extracting features, and it is necessary to study a more effective method for extracting features. In addition, it is known that the authentication rate of pinna images is significantly reduced when there are obstructions to the auricle such as hair, ornaments such as piercings, or when the light level is not sufficient. In these situations, features based on the PRTF may be superior to those based on pinna images, and should be considered for practical use.\par
In addition, the use of time-series data of acceleration and angular velocity for learning can be considered. In this study, we used the acceleration information at the time when the smartphone was held against the ear, but we believe that using the time-series data of acceleration and angular velocity for learning will improve the robustness of the personal authentication system.

\ifCLASSOPTIONcompsoc
  \section*{Acknowledgments}
\else
  \section*{Acknowledgment}
\fi
This study is financially supported by JSPS KAKENHI (18K19791).

\ifCLASSOPTIONcaptionsoff
  \newpage
\fi



%

%

\begin{IEEEbiography}
[{\includegraphics[width=1in,height=1.25in,clip,keepaspectratio]{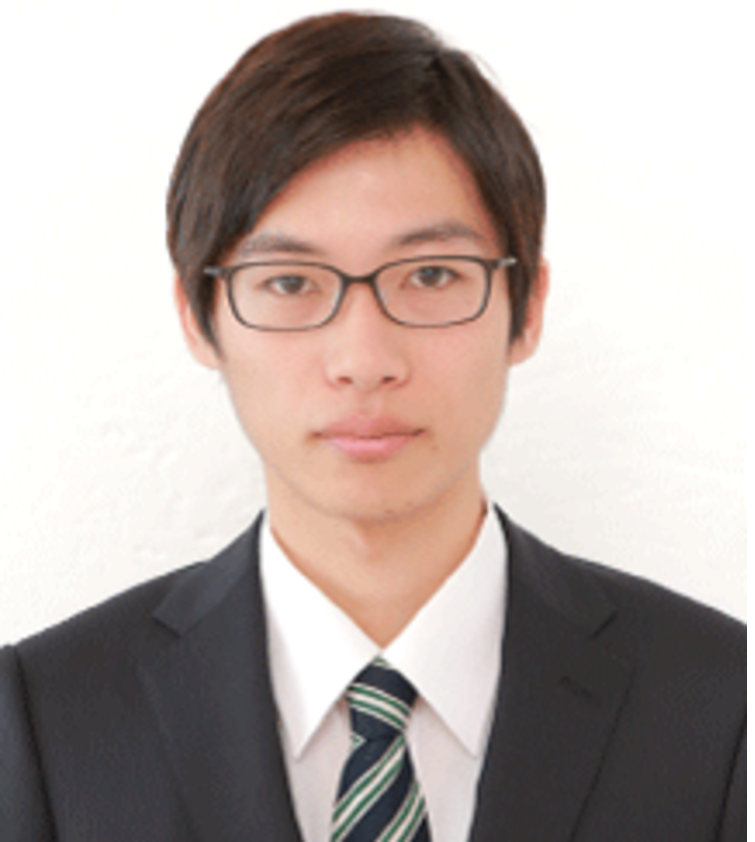}}]{Shunji Itani}

received the B.E. degree from Kansai University, Osaka, Japan in 2019. He is now a master course student of Kansai University Graduate School. His research interest is machine learning, signal processing, and personal authentication. 
\end{IEEEbiography}



\begin{IEEEbiography}
[{\includegraphics[width=1in,height=1.25in,clip,keepaspectratio]{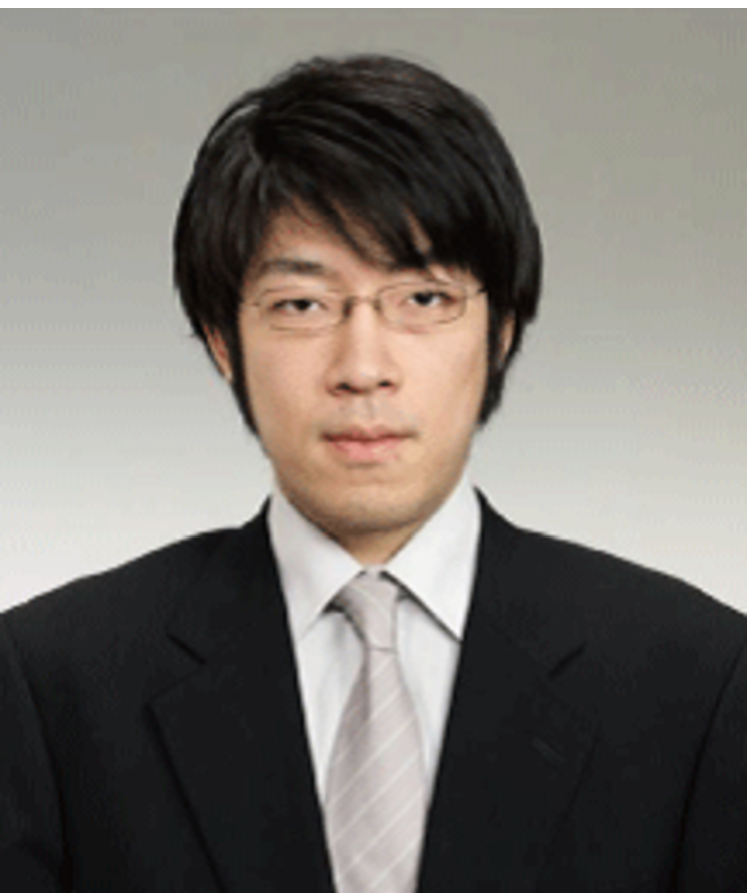}}]{Shunsuke Kita}
received the M.E. degree from Kansai University, Osaka, Japan in 2011.
He is now Osaka Research Institute of Industrial Science and Technology (ORIST), the research division of electronic and mechanical systems.
His research interests include machine-learning and acoustic-structural coupled analytical methods.
\end{IEEEbiography}

\begin{IEEEbiography}
[{\includegraphics[width=1in,height=1.25in,clip,keepaspectratio]{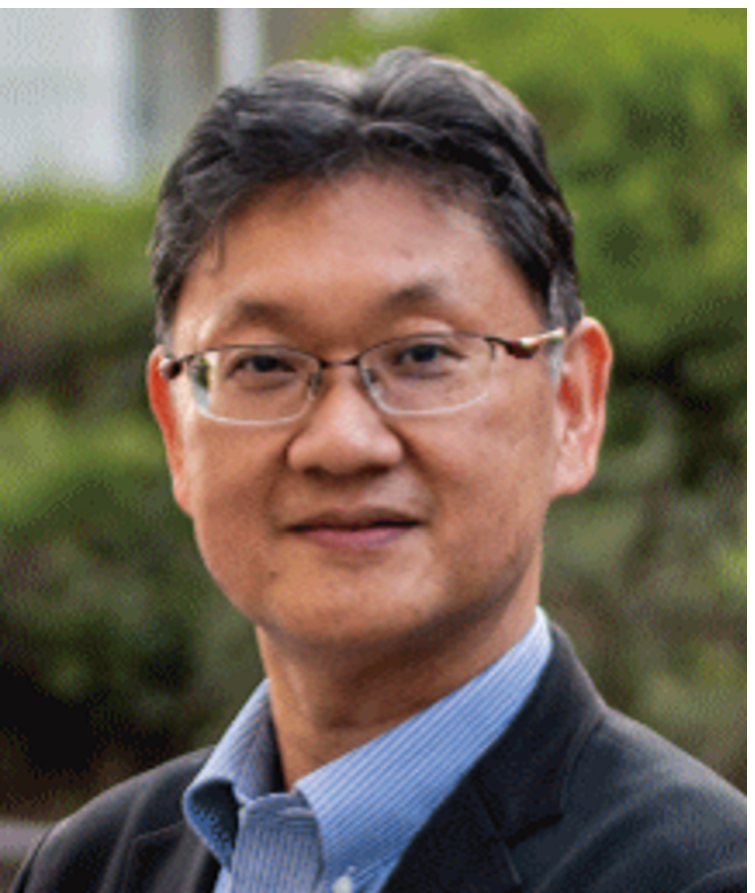}}]{Yoshinobu Kajikawa}
received his B.Eng. and M.Eng. degrees in electrical engineering from Kansai University, Osaka, Japan, in 1991 and 1993, respectively. He received his D.Eng. degree in communication engineering from Osaka University, Osaka, Japan, in 1997. He joined Fujitsu Ltd., Kawasaki, Japan, in 1993, and engaged in research on active noise control. In 1994, he joined Kansai University, Osaka, Japan, where he is currently a professor. His current research interests include signal processing for audio and acoustic systems. He has authored or coauthored more than 200 articles in journals and conference proceedings and has more than 10 patents. He received the 2012 Sato Prize Paper Award from the Acoustical Society of Japan (ASJ), the Best Paper Award in APCCAS 2014, the 2017 Sadaoki Furui Prize Paper Award from the Asia Pacific Signal and Information Processing Association (APSIPA), and the 2019 Best Paper Award from IEICE. He is a senior member of IEEE and IEICE, and a member of APSIPA, Acoustical Society of America (ASA), and ASJ. He is now serving as the Editor-in-Chief of IEICE Transactions on Fundamentals of Electronics, Communications, and Computer Sciences. He is also an associate editor of the IET Signal Processing and Applied Sciences, and a member of SLA TC in APSIPA. He is now a vice-president of APSIPA and a vice-chair of TC on Engineering Acoustics in IEICE. He was a vice-president of IEICE Engineering and Science Society.
\end{IEEEbiography}



\end{document}